\documentclass[lettersize,journal]{IEEEtran}
\IEEEoverridecommandlockouts  
\usepackage{color}
\usepackage{amsmath,amsfonts}
\usepackage{algorithmic}
\usepackage{algorithm}
\usepackage{array}
\usepackage[caption=false]{subfig}
\usepackage{tabularx}
\usepackage{textcomp}
\usepackage{vcell}
\usepackage{stfloats}
\usepackage{multirow}
\usepackage{ulem}

\usepackage{verbatim}
\usepackage{graphicx}
\usepackage{cite}
\usepackage{amsmath,amsfonts}

\usepackage{flushend}
\usepackage{diagbox}

\usepackage{array}
\usepackage[caption=false]{subfig}
\usepackage{textcomp}
\usepackage{color}
\usepackage{stfloats}
\usepackage{url}
\usepackage{verbatim}
\usepackage{graphicx}
\usepackage{soul}
\sethlcolor{yellow}
\usepackage{cite}
\usepackage{tabularx}
\usepackage[flushleft]{threeparttable}
\usepackage{booktabs}
\usepackage{empheq}
\usepackage{makecell}
\usepackage[dvipsnames]{xcolor}
\usepackage{xcolor}

\usepackage{tikz}

\newcommand\copyrighttext{%
  \footnotesize This is the author’s version of the work. It is posted here by permission of IEEE for personal use, not for redistribution. The definitive version was published in IEEE Transactions on Medical Robotics and Bionics, doi: 10.1109/TMRB.2026.3699216.}
\newcommand\copyrightnotice{%
\begin{tikzpicture}[remember picture,overlay]
\node[anchor=south,yshift=10pt] at (current page.south) {\fbox{\parbox{\dimexpr\textwidth-\fboxsep-\fboxrule\relax}{\copyrighttext}}};
\end{tikzpicture}%
}

\begin{document}
\title{Design and Evaluation of a Compliant Quasi-Direct Drive End-Effector for Safe Robotic Ultrasound Imaging

\thanks{This work is supported in part by the National Science Foundation Traineeship in the Advancement of Surgical Technologies (TAST) program under Grant 2125528, and the Innovation Co-Lab at Duke University’s Office of Information Technology 

Danyi Chen and Brian P. Mann are with the Dynamical Systems Lab, Thomas Lord Department of Mechanical Engineering and Materials Science, Duke University, NC, USA

Ravi Prakash, Vincent Y. Wang, Zacharias Chen, Sarah Dias, Leila J. Bridgeman, and Siobhan R. Oca are with the Thomas Lord Department of Mechanical Engineering and Materials Science, Duke University, NC, USA

Daniel M. Buckland is with Emergency Medicine \& Mechanical Engineering Department, University of Wisconsin-Madison, WI, USA

$^\dagger$Corresponding author: will.chen@duke.edu; 
}
}

\author{Danyi Chen$^\dagger$, Ravi Prakash, Vincent Y. Wang, Zacharias Chen, Sarah Dias, Daniel M. Buckland, \\Leila J. Bridgeman, Siobhan R. Oca, and Brian P. Mann}

\maketitle
\copyrightnotice
\begin{abstract}

Robot-assisted ultrasound scanning promises to advance autonomous and accessible medical imaging. However, ensuring patient safety and compliant human-robot interaction during probe contact poses a significant challenge. Most existing systems either have high mechanical stiffness or trade performance for compliance. This paper presents a novel compliant end-effector designed to mount on robotic arms for safe and accurate robotic ultrasound imaging, using a quasi-direct drive actuator to achieve passive mechanical compliance and precise active force control. To evaluate the end-effector's performance, we developed an \textit{ex vivo} dynamic motion simulator platform for contact and scanning testing on tissue under simulated movements. The end-effector was evaluated against a UR3e robot arm using conventional force control strategies as a baseline. Single-point contact experiments from 2.5 N to 15 N show that the end-effector reduced force tracking RMS error by  80.1\% on average. Trajectory scanning experiments at different speeds showed an average of 68.0\% reduction in force tracking error. Statistically significant improvements were observed in four of six quantitative image quality and stability metrics using the end-effector. This work presents a novel approach for designing and evaluating compliant end-effectors, with the goal of improving safety and reliability in robotic ultrasound.
\end{abstract}


\section{Introduction}
\IEEEPARstart{U}{ltrasound} (US) imaging has been a reliable and versatile clinical tool since the mid-20th century, widely used due to its non-invasive nature, real-time imaging, lack of ionizing radiation, and relatively low cost \cite{fischerova_ultrasound_2011}. However, conventional US exams rely on trained sonographers to manually apply consistent pressure and precisely control the probe's orientation.  Between 2011 and 2021, the growth in US examinations (55.1\%) outpaced the rise in the number of sonographers (43.6\%), especially in remote or underserved areas\cite{won2024sound,nathan_evaluation_2017}. The United States Bureau of Labor Statistics projects a need for an additional 27,600 sonographers by 2024\cite{noauthor_current_nodate}. In the United Kingdom, 2023 data showed about 416,900 patients were waiting 6 weeks or more for key diagnostic tests that include US, with one-third of patients waiting more than 6 weeks for diagnostic tests requiring non-obstetric US\cite{noauthor_sor_nodate}. Robotic US systems could offer significant advantages by automating probe manipulation and improving imaging precision and reproducibility \cite{li_overview_2021}. Furthermore, robotic US systems can also operate autonomously, enabling US imaging in high-latency or isolated environments such as space missions, remote locations with limited healthcare professionals, or hyperbaric chambers used in diving operations where direct medical supervision may be restricted \cite{li_overview_2021}.
\begin{figure}[h]
    \centering
    \includegraphics[width=0.5\textwidth]{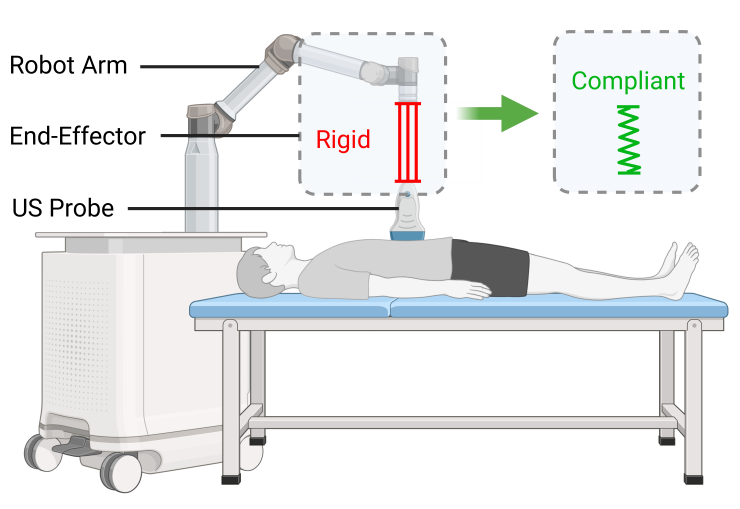} 
    \captionsetup{justification=justified}
    \caption{Schematic diagram of robotic US imaging components, showing a robotic arm with mounted US probe positioned over a patient on an examination table. Because the robotic probe contacts the patient directly, the choice between rigid and compliant end-effectors matters for safe imaging.\cite{prakash2025biorender}.}
    \label{fig:demo}
\end{figure}

The safe integration of robotic US systems in clinical practice depends on achieving proper contact compliance: the robot's ability to adapt and yield to patient movements without exerting excessive pressure. A major concern is the potential for excessive force exerted by the robot end-effector, typically rigid and non-mechanically compliant, which can lead to patient discomfort or even injury \cite{kuo_automatic_2023, wang_compliant_2023}. This is especially important for abdominal or retroperitoneal imaging, where instinctive movements of the body, such as respiration and tensing in response to touch, would require a robotic system to adapt dynamically to maintain consistent and safe contact with the skin \cite{virga_automatic_2016}. While patients can hold their breath, this causes delays in scanning sessions that can last up to 30 minutes. Autonomous systems that can adapt to normal patient movement could shorten scans and be more comfortable. 

Researchers have explored active and passive strategies to address compliance challenges \cite{wang_passive_1998}. Active compliance enables a rigid robot arm to exhibit compliant behavior by controlling the end-effector position based on force feedback to achieve compliant scanning through control algorithms alone. State-of-the-art approaches often use task-space force controllers such as admittance or PID controllers to simulate compliant behaviors \cite{mathiassen_ultrasound_2016, fang_force-assisted_2017, wang2022full, rigby2022development}. Advanced impedance control approaches have also demonstrated effective force regulation on scanning surfaces~\cite{dyck2022impedance,suligoj2021robust}. However, these methods are limited by lower bandwidths, making them slower to respond to sudden movements \cite{chignoli_mit_2021}. In addition, most commercial rigid robot arms only support position or velocity control modes. This restriction forces compliance to be implemented at a higher control level through position or velocity corrections, resulting in slower force response times and reduced stability compared to systems with native current or torque control capabilities. Passive compliance, achieved by integrating mechanically compliant elements, allows the robot to physically deflect or deform in response to external forces without a controller. This includes soft materials, elastic actuators, mechanical clutches, and pneumatic actuators\cite{lindenroth_design_2017,wang2019bespoke,kuo_automatic_2023,bao2021constant,wu2025robotic}. However, each approach has significant limitations. Soft end-effectors can apply variable force based on deformation, while linear elastic actuators have very limited operational position range due to the physical constraints of spring deflection limits~\cite{bao2021constant, wu2025robotic}. Pneumatic actuators can regulate force but are nonlinear, with limited control bandwidth~\cite{zoller_acoustic_2018, das_incremental_2021}. Additionally, some approaches~\cite{lindenroth_design_2017,kuo_automatic_2023} lack position measurement capability, which is needed for 3D reconstruction.

\begin{table*}[htbp]
  \centering
  \caption{Comparative overview of state-of-the-art compliant robotic US systems and our proposed system}
  \begin{tabularx}{\textwidth}{l*{5}{>{\centering\arraybackslash}X}}
    \toprule
    & \textbf{Operating Force Range$^e$} & \textbf{Operating Position Range} & \textbf{Feedback Types} & \textbf{Active Force Regulation} & \textbf{Mechanical Compliance} \\
    \midrule
    \textbf{Our Proposed System} & \textbf{2.5 N - 15 N} & \textbf{52 mm} & \textbf{Position, Force} & \textbf{Yes} & \textbf{Yes} \\
    Wu25$^c$\cite{wu2025robotic} & 12.3 N - 21.3 N & 3 mm & Position, Force & Yes & Yes \\
    Wang23$^a$\cite{wang2022full} & 3 N - 15 N & N/A & Position, Force & Yes & No \\
    Kuo23$^d$\cite{kuo_automatic_2023} & 4.1 N - 7.5 N & 50 mm & Force & Yes & Yes \\
    Bao21$^c$\cite{bao2021constant} & 4 N - 24 N & 8 mm & Position, Force & Yes & Yes \\
    Suligoj21$^a$\cite{suligoj2021robust} & 1.6 N - 11.9 N & N/A & Position, Force & Yes & No \\
    Lindenroth17$^b$\cite{lindenroth_design_2017} & 0 N - 10 N & 6.3 mm & None & No & Yes \\
    Mathiassen16$^a$\cite{mathiassen_ultrasound_2016} & 1 N - 5 N & N/A & Position, Force & Yes & No \\
    
    \bottomrule
  \end{tabularx}%
  
  \vspace{0.5em}
  \footnotesize
  $^a$Robot arm only systems; $^b$Soft actuator; $^c$Elastic actuator; $^d$Pneumatic actuator.
  \\
  $^e$Clinical requirements for force: 7.5 N mean, 17.3 N max \cite{dhyani_precise_2014}
  \\
  Pain/discomfort threshold lower-bound: 20.0 N (extrapolated) \cite{waller2025pressure}
  
  \label{tab:tab1}%
\end{table*}%

While electric motors offer improved bandwidth and ease of control compared to pneumatic and fluidic actuators, traditional electric motors used in robotic arms often lack mechanical compliance due to their high gear ratios and significant backdrive force \cite{kakogawa_highly_2022}. However, recent advancements have led to the development of quasi-direct drive (QDD) actuators, which can achieve high torque control fidelity while maintaining low backdrive torque due to their low gear ratios. This design allows external forces to backdrive the motor and achieve compliance mechanically \cite{singh_design_2020}. 

For evaluation, a natural human movement mimicking platform is needed to perform robust and realistic evaluation of compliant US systems. While several approaches have been developed to ensure safe and effective force control during US procedures\cite{mathiassen_ultrasound_2016, fang_force-assisted_2017, wang2022full, rigby2022development,lindenroth_design_2017,kuo_automatic_2023,bao2021constant,wu2025robotic}, existing testing tissue platforms are predominantly static. As a result, current tissue platforms have limited capability to accurately replicate the dynamic behavior of both physiological and unexpected human motion during abdominal or retroperitoneal US exams. This gap in available testing infrastructure restricts the ability to fully evaluate the performance of compliant US systems in realistic scenarios.

This paper presents two main contributions. The first is the mechanical design and dynamic modeling of a novel single-degree-of-freedom compliant end-effector for robotic ultrasound imaging, using QDD actuation to provide passive mechanical compliance for safe patient interaction, precise active force control for real-time response to dynamic movements, and spatial feedback of probe position across a large operating range. The second is an actuated \textit{ex vivo} tissue platform that simulates abdominal and back motion during breathing and sudden unexpected movements, providing a realistic and repeatable evaluation methodology applicable to robotic ultrasound systems beyond the proposed end-effector.

\section{Methods}

\subsection{Clinically Informed Design Requirements}

Clinical US imaging requirements inform our design specifications, focusing on key metrics often inadequately addressed by existing solutions: operating position and force range, active force control, probe position feedback, and mechanical compliance. Force capability presents a particular challenge. For normal weight patients, sonographers typically apply contact forces of 7.5 N (mean) and 17.3 N (maximum) during US scans \cite{dhyani_precise_2014}. However, patient comfort limits must also be considered. Waller et al. \cite{waller2025pressure} reported pressure pain threshold reference values in pain-free older adults, with the lowest recorded threshold of 75 kPa. When extrapolated to the US transducer's contact surface used in this study (Interson SP-L01, Interson Corporation, Pleasanton, CA) of 38 mm by 7 mm, this threshold corresponds to approximately 20.0 N of applied force. As shown in Table \ref{tab:tab1}, some state-of-the-art compliant US end-effectors fall short of reaching the clinically required force while remaining within patient comfort limits.

\subsection{High-Performance Compliant Actuation}
\noindent

Achieving both passive compliance and precise active force control is a fundamental challenge in robotic actuation. While various actuation approaches have been used in state-of-the-art solutions, each comes with trade-offs. This section compares these established actuation methods and shows why we selected QDD actuators, which address the limitations of existing approaches while maintaining the dual requirements of compliance and control fidelity.

Compared to pneumatic actuators, QDD systems eliminate complex pressure regulation systems, which are often nonlinear and exhibit slower response times \cite{zoller_acoustic_2018, das_incremental_2021}. Additionally, QDD actuators provide built-in positional feedback\cite{katz_mini_2019}, eliminating the need for external position sensors.

Compared to elastic actuators, QDD actuators avoid the fundamental trade-off between control fidelity and compliance. While elastic actuators must balance these characteristics through spring elements, QDD actuators achieve both high control fidelity and backdrivability simultaneously through their low gear ratio design \cite{Junior16}. This approach also reduces system complexity by eliminating springs and torque sensors, as torque can be estimated accurately using motor current\cite{10611567}.

Compared to soft fluidic actuators, QDD systems provide more predictable behavior and simplified control strategies, avoiding the unpredictable deformation and complex feedback systems in soft systems \cite{li2022soft}.

QDD actuators have been successfully implemented in various fields requiring compliance, safe human-robot-interaction, and impact absorption, including powered prostheses\cite{azocar_design_2020}, powered orthoses\cite{long_design_2022}, and legged robotics\cite{chignoli_mit_2021, katz_mini_2019}. For this study, we chose the AK60-6 (Cubemars, Shenzhen, China) QDD actuator, with a rated torque of 3 Nm, a backdrive torque of 0.2 Nm, and built-in positional feedback.

\subsection{Transmission Selection for Optimal Control and Safety}

\begin{figure}[h!]
    \centering
    \includegraphics[width=0.49\textwidth]{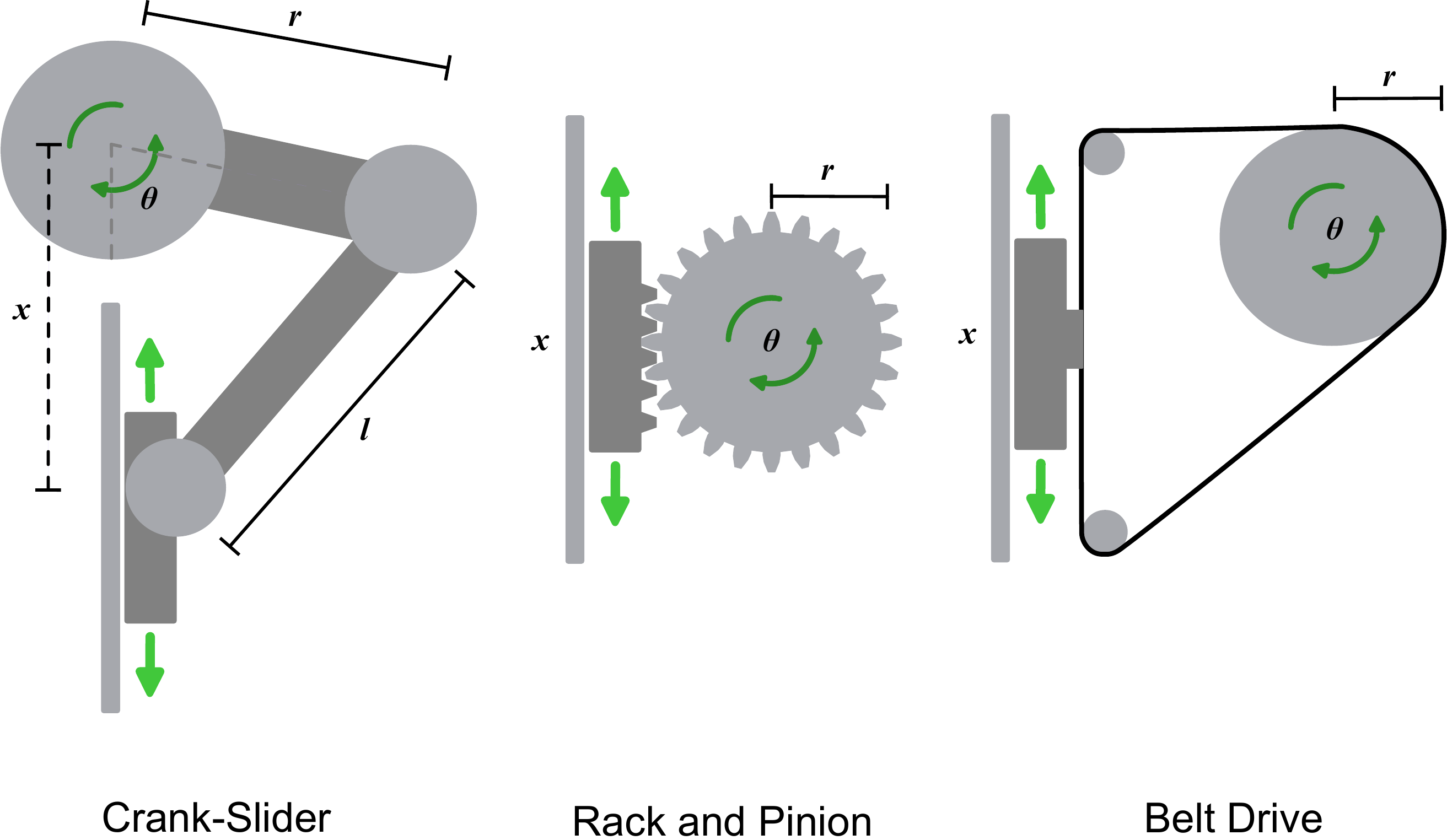} 
    \captionsetup{justification=justified}
    \caption{\small Comparative analysis of three mechanical transmissions for rotational-to-linear motion conversion. From left to right: crank-slider mechanism, rack and pinion gear system, and belt drive system. Each mechanism is illustrated with its kinematic configuration (green arrows indicate direction of motion) and operational parameters, the section shows the belt drive's advantages of linear motion transmission and reflected inertia, low friction, and inherent compliance for robotic US applications.}
    \label{fig:mechanism}
\end{figure}

Converting the QDD’s rotational motion into linear motion requires a transmission that preserves the actuator’s strengths in backdrivability and control precision, while complementing the safety/compliance capabilities. This section describes the primary design considerations and compares multiple transmission types we evaluated. The criteria we evaluated were control linearity, reflected inertia, friction characteristics, and compliance. Reflected inertia and friction in particular contribute to the backdrive performance which is critical to complement the QDD actuator. We eliminated lead screws (poor backdrivability) and ball screws (excessive cost), then evaluated three viable candidates: crank-slider, rack and pinion, and timing belt driven systems, as shown in Fig. \ref{fig:mechanism}. The belt-driven transmission ultimately met our requirements best, maintaining backdrivability while eliminating the backlash and kinematic limitations present in the other mechanisms.

\subsubsection{Crank-Slider}
Our first prototype used the crank-slider mechanism, but we ultimately transitioned away from it due to its limitations. While capable of high force transmission, the crank-slider exhibits nonlinear motion characteristics with variations throughout its stroke. By the principle of virtual work, force transmission is the inverse of displacement transmission, so this nonlinearity can make controllers more challenging to design, and can be understood through its kinematic equation
\begin{equation}
x = r\cos\theta + \sqrt{l^2 - r^2\sin^2\theta}
\end{equation}
where $x$ is the linear displacement, $r$ is the length of the crank, $l$ is the length of the connecting rod, and $\theta$ is the angle of rotation. The derivative of this relationship
\begin{equation}
\frac{dx}{d\theta} = -r\sin\theta - \frac{r^2\sin\theta\cos\theta}{\sqrt{l^2 - r^2\sin^2\theta}} \quad \text{(variable)}
\end{equation}
reveals that the rate of change of the linear displacement varies nonlinearly with $\theta$, thus creating control challenges, especially at the extremes of motion where mechanical advantage approaches zero.

The reflected inertia for the crank-slider mechanism

\begin{equation}
m_{\text{eff}} = \frac{J_{m} N^2}{J(\theta)^2}
\end{equation}
where $J_{m}$ is the motor rotor inertia, $N$ is the gear ratio = 6 for our actuator, and $J(\theta) = \frac{dx}{d\theta}$ is the kinematic Jacobian of the transmission. This shows that when approaching the dead points ($\theta = 0$ and $\theta = \pi$), the Jacobian $J$ approaches zero, causing the reflected inertia to approach infinity. Furthermore, the position-dependent nature of the reflected inertia complicates force control throughout the workspace.

Friction within the crank-slider mechanism comes from multiple sources: the crank pin joint, the wrist pin joint, and the slider-guide interface. However, all three locations can use rolling element bearings, reducing friction losses compared to sliding contacts.

Lastly, the rigid nature of the system lacks any inherent mechanical compliance, reducing the tolerance for safe human-robot-interaction.

\subsubsection{Rack and Pinion}
The rack and pinion system offers excellent positioning accuracy and a direct linear relationship between input rotation and output motion.

The kinematic equation is:
\begin{equation}
x = r \theta
\end{equation}
where $r$ is the pitch radius of the pinion, yielding a constant Jacobian:
\begin{equation}
J = \frac{dx}{d\theta} = r .
\end{equation}

The reflected inertia is constant throughout the motion:
\begin{equation}
m_{\text{eff}} = \frac{J_m N^2}{r^2}
\end{equation}
This eliminates the nonlinearity present in the crank-slider mechanism.

\begin{figure*}[ht!]
    \centering
    \includegraphics[width=\textwidth]{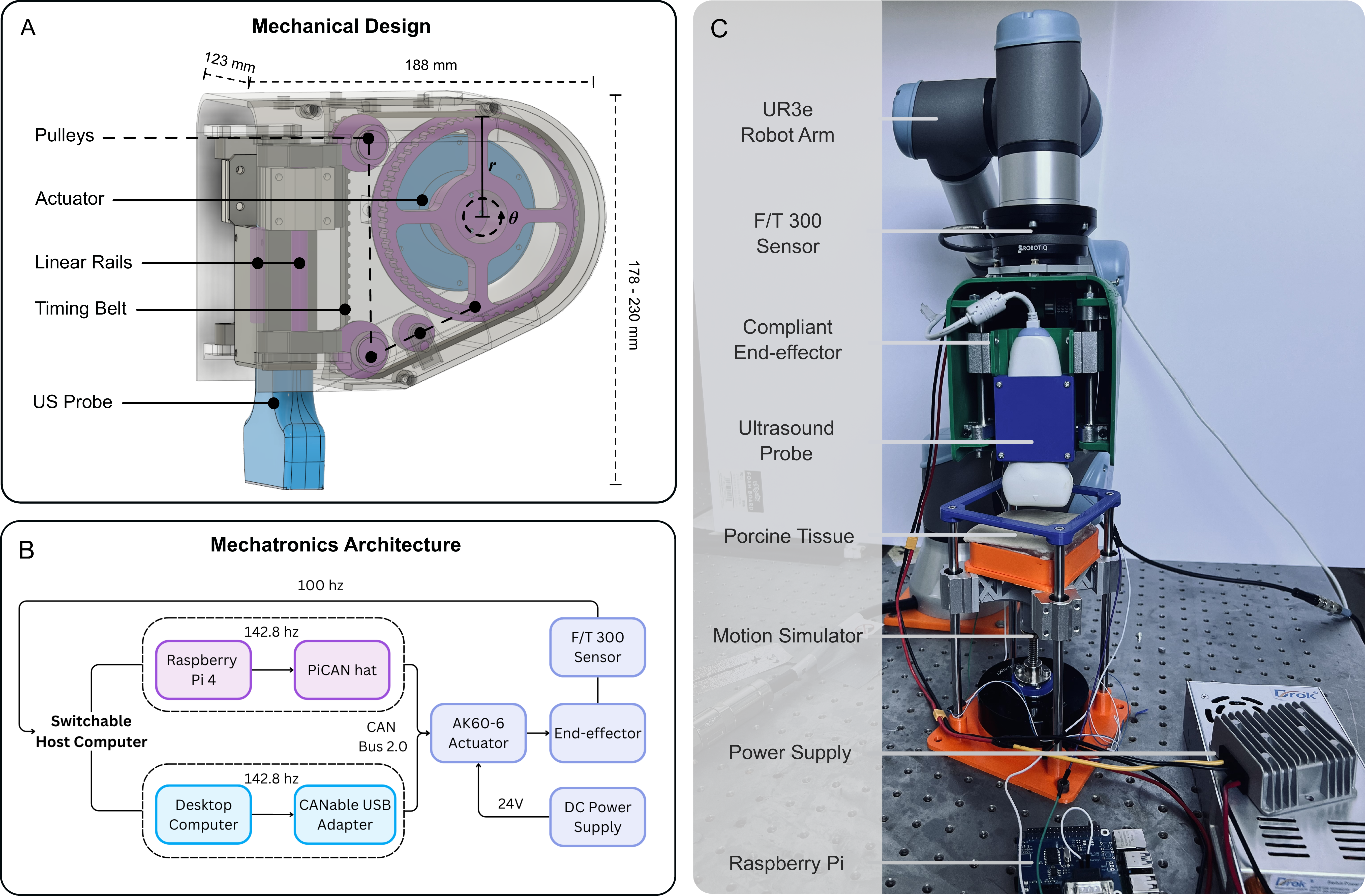} 
    \captionsetup{justification=justified}
    \caption{\small Overview of the mechatronic system design and setup, demonstrating the complete development pipeline from mechanical design through electronics architecture to experimental validation. A: schematic diagram breakdown of the compliant robotic US end-effector. The QDD actuator drives a large-diameter pulley with radius $r$. The rotational angle $\theta$ translates to linear motion via a timing belt driving the US probe mount on linear rails, a tensioner pulley is used to maintain belt tension; B: mechatronics architecture of the end-effector. The mechatronics setup also powers the dynamic motion simulator; C: experiment setup of the end-effector attached to the UR3e robot arm in the point contact setup, enabling evaluation of the system under realistic conditions.}
    \label{fig:diag}
\end{figure*}

Despite the favorable inertia characteristics, the rack and pinion mechanism suffers from significant friction due to the sliding contact between gear teeth during meshing. This gear mesh friction can decrease the backdrivability. Additionally, the gear interface may introduce backlash, which affects control precision and can cause force discontinuities during direction reversals.

Like the crank-slider mechanism this rigid mechanical coupling transmits all impact forces directly to the actuator and like the crank-slider mechanism, also lacks inherent mechanical compliance. In addition, the linear input-output and reflected inertia characteristics are not unique to this transmission, as belt-driven systems also exhibit this behavior.

\subsubsection{Belt-Driven Transmission}
In contrast, the belt-driven transmission system provides several distinct advantages. Like the rack and pinion system, the timing belt system has a simpler force transmission compared to the crank-slider linkages, resulting in a more linear relationship, as described by:
\begin{equation}
x = r\theta
\end{equation}
where $x$ is the linear displacement, $r$ is the radius of the pulley, and $\theta$ is the angle of rotation. The derivative,
\begin{equation}
\frac{dx}{d\theta} = r \quad \text{(constant)}.
\end{equation}
shows a constant rate of change, resulting in a direct linear relationship between input rotation and output motion.

The reflected inertia for the belt-driven system is:
\begin{equation}
m_{\text{eff}} = \frac{J_m N^2}{r^2} .
\end{equation}
Like the rack and pinion, the belt system maintains constant reflected inertia throughout its entire range of motion, with no singularities or position-dependent variations.

However, the belt-driven system has minimal friction losses due to its reliance on rolling contact elements: the pulley rotates on ball bearings, and the probe carriage travels on linear rail bearings. Unlike the rack and pinion, there is no sliding friction in the primary transmission path. This low-friction design preserves the QDD's inherent backdrivability, enabling compliance and safe patient interaction.

Beyond this, the inherent properties of the belt material provide intrinsic impact absorption, adding compliance to the system over the other methods. This belt compliance changes the system dynamics: while rigid transmission systems (rack and pinion and crank-slider) exhibit second-order dynamics, the belt's elasticity introduces fourth-order dynamics that naturally absorbs impacts, which will be explored further in the dynamic model section when formulating the dynamic equations. The belt system also requires relaxed manufacturing tolerances than the other two systems and provides natural overload protection through controlled slippage. While the belt dynamics does introduce some nonlinearity, the linear kinematics relation and reflected inertia, as well as superior friction and compliance characteristics of the belt drive make it well-suited for this application.

\subsection{Design}
The complete design is illustrated in Fig. \ref{fig:diag}A. The actuator drives a 96.5 mm diameter pulley, which is coupled to a linear slide mechanism housing the US probe. The belt is tensioned using an adjustable tensioning pulley integrated into the frame design. The design also uses belt slippage as a passive overload protection mechanism. When contact forces generate motor torques sufficient to slightly deform the frame structure, the belt tension reduces enough to allow slippage on the pulleys, preventing force transmission above a threshold determined by the frame stiffness (experimentally, the prototype exhibited slippage at forces $>$ 15 N). The probe has a maximum travel distance of 52 mm, sufficient to accommodate abdominal movement even during deep breathing \cite{kaneko2012breathing}. The housing is fabricated from PLA plastic. With a total weight of 1.13 kg and a length ranging 178 - 230 mm (including the US probe) along the tool Z-axis depending on the extension. When fully retracted, the end-effector adds just 36 mm to the 142 mm Interson SP-L01 probe. The system is compatible even with small robot arms such as the UR3e, which has a maximum payload of 3 kg.  This compact form factor makes navigation around the human body easier compared to longer end-effectors. While QDDs typically have larger footprints that may require self-collision planning for complex motions, we designed the end-effector's geometry to concentrate most of its volume offset to one side (Fig. \ref{fig:diag}A), allowing the bulk to be angled away from the robot arm during simple scanning trajectories like the zigzag motion demonstrated in this work.

A six-axis Robotiq F/T 300 sensor (Robotiq, Lévis, QC, Canada) is mounted between the robot arm and the end-effector, directly along the z axis of the US probe to provide force feedback and validation. While the QDD actuator provides current-based torque estimation, the external F/T sensor offers a widely accepted benchmark for force control evaluation. A limitation of this configuration is that the sensor may measure aggregate forces including end-effector inertia. The backdrive force, which represents the minimum external force required to move the probe when the system is not powered, is calculated to be 4.1 N.

\subsection{Dynamic Model of the QDD End-Effector System}
To understand how forces and motions propagate through the system, we need a mathematical model of the  actuator-transmission-probe 
chain. This section derives an idealized frictionless model to analyze the fourth-order dynamics of the QDD actuator and belt system, which introduces compliance beyond the second-order behavior of rigid transmissions.

The end-effector is governed by the coupled dynamics of the QDD actuator and timing belt transmission. The motor dynamics follow the fundamental torque balance equation,
\begin{equation}
J_m \ddot{\theta} + b_m \dot{\theta} = K_t I - \frac{\tau_l}{N}
\end{equation}
obtained from the manufacturer's datasheet, where $J_m = 2.435 \times 10^{-5} $kg$\cdot$m$^2$ is the motor inertia, $b_m = J_m/\tau_{\text{mech}} = 0.03$ Nm$\cdot$s/rad is the viscous damping coefficient derived from the mechanical time constant $\tau_{\text{mech}} = 0.81$ ms, $K_t = 0.078$ Nm/A is the torque constant, $N = 6$ is the gear ratio, $I$ is the motor current, $\theta$ is the motor shaft angle, and $\tau_l$ is the load torque at the output shaft. The timing belt introduces compliance between the motor output and the linear probe motion,
\begin{equation}
F_{\text{belt}} = k_b(\frac{r\theta}{N} - x)
\end{equation}
where $k_b = 2013.8$ N/m is the nominal stiffness of a GT2 timing belt under moderate (low-strain) tension \cite{robotics7040075}, $r = 0.04825$ m is the pulley radius, and $x$ is the probe's true linear displacement. Applying Newton's second law to the probe assembly yields
\begin{equation}
m_{\text{eff}}\ddot{x} = F_{\text{belt}} - F_{\text{contact}} + F_g
\end{equation}
where $m_{\text{eff}} = m_{\text{probe}} + \frac{J_m N^2}{r^2}$ is the effective mass including the reflected motor inertia, $F_{\text{contact}}$ is the tissue interaction force measured by the force sensor, and $F_g = m_{\text{probe}}g\cos\alpha$ is the gravitational force component along the probe axis, with $\alpha$ being the angle between the probe axis and the vertical direction.

Combining these equations yields the complete system dynamics. Defining the state vector $\mathbf{x} = [\theta, \dot{\theta}, x, \dot{x}]^T$ and inputs $\mathbf{u} = [I, F_{\text{contact}}]^T$ (where $I$ is the control current input and $F_{\text{contact}}$ is the measured contact force), the system can be written in linear state-space form
\begin{equation}
\dot{\mathbf{x}} = \mathbf{A}\mathbf{x} + \mathbf{B}\mathbf{u} + \mathbf{c}
\end{equation}
where:
\begin{align}
\mathbf{A} &= \begin{bmatrix}
0 & 1 & 0 & 0 \\
-\frac{r^2k_b}{J_mN^2} & -\frac{b_m}{J_m} & \frac{rk_b}{J_mN} & 0 \\
0 & 0 & 0 & 1 \\
\frac{rk_b}{m_{\text{eff}}N} & 0 & -\frac{k_b}{m_{\text{eff}}} & 0
\end{bmatrix} \\
\mathbf{B} &= \begin{bmatrix}
0 & 0 \\
\frac{K_t}{J_m} & 0 \\
0 & 0 \\
0 & -\frac{1}{m_{\text{eff}}}
\end{bmatrix}, \quad
\mathbf{c} = \begin{bmatrix}
0 \\
0 \\
0 \\
\frac{F_g}{m_{\text{eff}}}
\end{bmatrix}.
\end{align}
This linear fourth-order model describes the dynamic model of the compliant end-effector. The frictionless dynamic model is derived but not used in the controller to isolate the mechanical effects during the main evaluations; as a result, the derived model was not experimentally validated as a part of this work.

\subsection{Mechatronics Architecture}

The mechatronics system architecture (Fig. \ref{fig:diag}) supports two host computer configurations: (1) a stationary desktop computer with a CANable USB-CAN adapter (Protofusion, Ellicott City, MD), or (2) a mobile Raspberry Pi 4 with a PiCAN2 hat (Copperhill Technologies, Greenfield, MA). Both configurations use CAN bus 2.0 for communication at speeds up to 1Mbps. A 48 V power supply feeds a buck converter, which steps down the voltage to 24 V to drive the AK60-6 QDD actuator. Force measurements are obtained using a six-axis F/T 300 force sensor, sampling at 100 Hz. The control loop executes at 142.8 Hz. While current implementations use stationary power sources, the Raspberry Pi-based system is designed to be easily adapted for mobile operation by incorporating a battery pack, making this system suitable for remote or resource-constrained environments. The system architecture also powers our dynamic motion simulator. Experiments were conducted on both the desktop computer and Raspberry Pi configurations.
\noindent

\subsection{Controller Design}
To evaluate the end-effector’s performance compared to a robot arm only baseline, we implemented simple controllers to isolate mechanical benefits while accounting for system-specific control input limitations (position, velocity, current). This section describes the three controllers used in our experiments.

\begin{figure}[h!]
    \centering
    \includegraphics[width=0.48\textwidth]{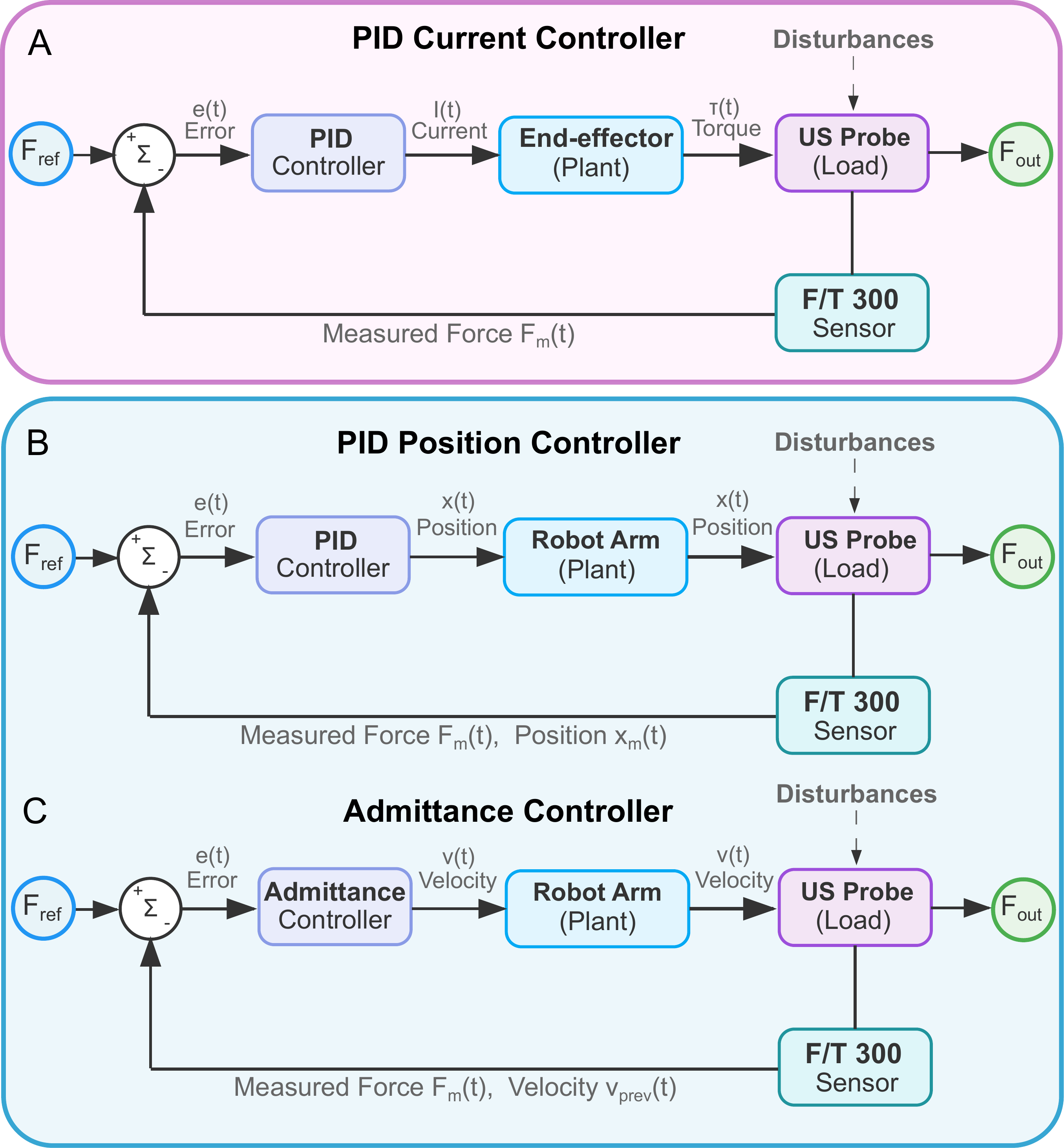} 
    \captionsetup{justification=justified}
    \caption{\small Controller architecture for both the end-effector and the robot arm. Only one controller is active during their respective experiments. A: the end-effector's current-based PID force controller; B: the robot arm's position-based PID force controller; C: the robot arm's admittance controller. These different control strategies provide comprehensive assessment of multiple methods of achieving compliant behavior.}
    \label{fig:cont}
\end{figure}

As illustrated in Fig. \ref{fig:cont}, Controller A is implemented on the end-effector, while Controllers B and C are implemented on the robot arm. Controller A and B are Proportional-Integral-Derivative (PID) controllers, a widely adopted approach\cite{zonetti2022pid}. Controller C is an admittance controller. 

The controller in Fig. \ref{fig:cont}A is a PID current controller implemented for the end-effector system. We chose to implement a simple PID controller instead of a model-informed controller using the dynamic model derived in Eq. 13-15 to better isolate the effects of mechanical design from control implementation. Direct current control is made possible by the mechanical design of the QDD actuator, which uses a low gear ratio that enables direct current-based force estimation and control. The low gear ratio minimizes mechanical impedance and allows for accurate force estimation through motor current measurements\cite{katz_mini_2019}.

The controllers in Fig. \ref{fig:cont}B-C were implemented on the robot arm platform as comparative baselines. During each experimental trial, only one of these controllers was active to ensure clear performance evaluation. The controller in Fig. \ref{fig:cont}B uses a PID position-based force control strategy, where force regulation is achieved by adjusting joint position commands. The controller uses position-based commands as commercial robot arms typically do not allow direct current control.  The controller implementation was based on the controller from \cite{rigby2022development}, and executes at 100 Hz. To optimize both PID controllers, we first used the Ziegler-Nichols tuning method as a starting point, and then tuned heuristically based on feedback. The gains were tuned until the best performance was observed for each condition. The gains used in the experiments are shown in Table \ref{tab:pid}.

\begin{table}[htbp]
  \centering
  \caption{PID gain values}
    \begin{tabularx}{0.5\textwidth}{cccccc}
    \toprule
    \textbf{} & $\textit{k}_{p}$ & $\textit{k}_{i}$ & $\textit{k}_{d}$ & \textbf{Force (N)} & \textbf{Experiment} \\
    \midrule
    \textbf{1}  & 0.629 & 8.85  & 0.015 & 2.5 & \multirow{4}{*}{\textbf{End-effector}} \\
    \textbf{2}  & 0.529 & 12.85 & 0.015 & 5 & \\
    \textbf{3}  & 0.429 & 8.85  & 0.015 & 10 & \\
    \textbf{4}  & 0.429 & 8.85  & 0.015 & 15 & \\
    \midrule
    \textbf{5}  & 0.03 & 5$\times$10$^{-8}$ & 0.001 & 2.5 & \multirow{4}{*}{\textbf{Robot arm only}} \\
    \textbf{6}  & 0.03 & 5$\times$10$^{-8}$ & 0.001 & 5 & \\
    \textbf{7} & 0.03 & 5$\times$10$^{-8}$ & 0.001 & 10 & \\
    \textbf{8} & 0.025 & 5$\times$10$^{-8}$ & 0.001 & 15 & \\
    \midrule
    \textbf{9$^a$}  & 0.26  & 2.39  & 0.01 & 5 & \multirow{2}{*}{\textbf{Sudden movement}} \\
    \textbf{10$^b$} & 0.02 & 5$\times$10$^{-8}$ & 0.001 & 5 & \\
    \bottomrule
    \end{tabularx}%
  \label{tab:pid}%
  \vspace{0.2cm}
  \footnotesize
  $^a$End-effector; $^b$robot arm only
\end{table}% 

The controller in Fig. \ref{fig:cont}C implements an admittance control architecture, which computes the desired velocities based on the measured force errors. Unlike the position-based approach, admittance controllers directly control the velocity. The controller implementation was based on the controller from\cite{wang2022full}, and executes at 20 Hz. 

\begin{figure*}[ht!]
    \centering
    \includegraphics[width=\textwidth]{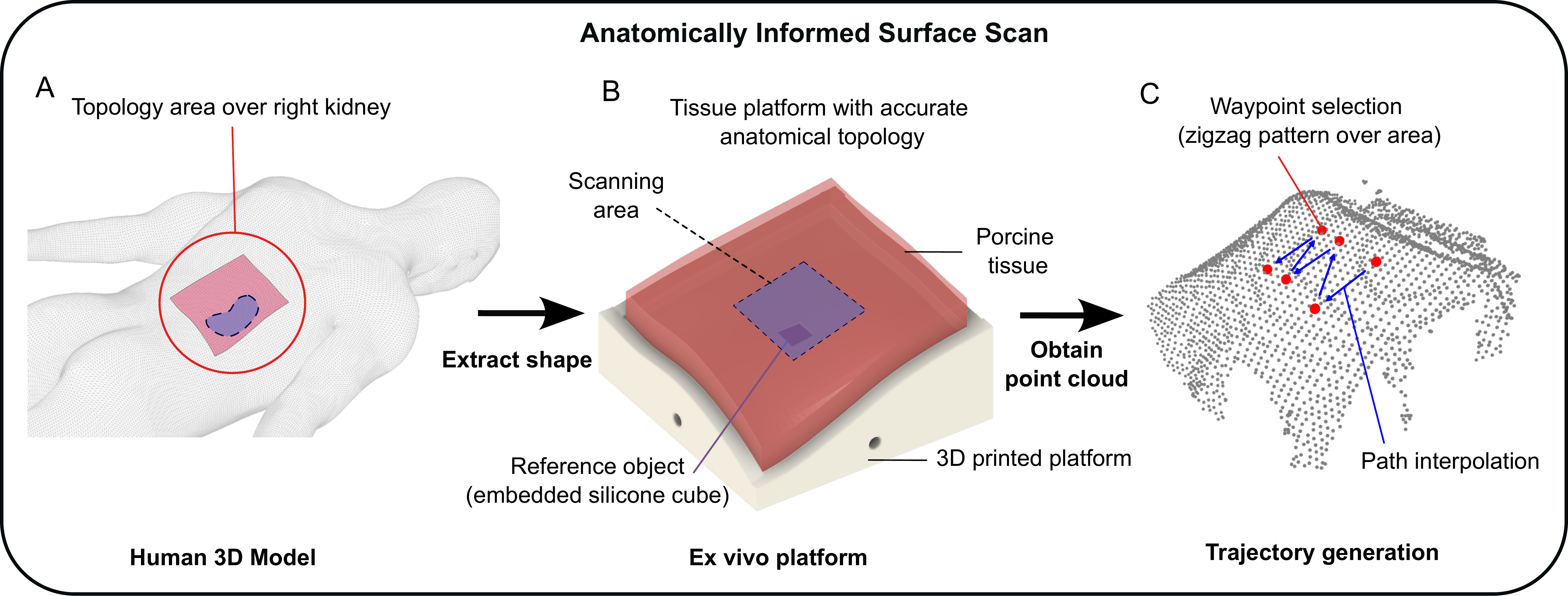} 
    \captionsetup{justification=justified}
    \caption{\small Anatomically informed testing platform development. A: human male 3D model with a highlighted posterior section above the right kidney; B: 3D-printed platform replicating uneven/non-flat topography of a human back, featuring a central silicone-filled cavity as US reference target, with porcine abdominal tissue on top; C: point cloud surface reconstruction obtained via depth camera scanning, with overlaid zigzag trajectory path for US probe navigation. This platform provides comprehensive evaluation by combining anatomically realistic, non-flat surface geometry with active motion during scanning, providing an evaluation environment under realistic conditions.}
    \label{fig:traj}
    
\end{figure*}

For the trajectory scan experiments, we implemented a path planning algorithm with a multi-segment trajectory generation scheme where the robot arm's tool link follows a sequence of waypoints $\mathbf{p}_i$ with corresponding surface normals $\mathbf{n}_i$. For each segment $i \rightarrow i+1$, the translational speed is computed as 
\begin{equation}
\mathbf{v}_{\text{trans}} = \frac{\mathbf{d}}{t_{\text{segment}}}
\end{equation}
where $\mathbf{d} = (\mathbf{p}_{i+1} + \mathbf{n}_{i+1} \ell) - (\mathbf{p}_i + \mathbf{n}_i \ell)$ represents the direction vector, $\ell$ is the probe length, and $t_{\text{segment}} = \frac{\|\mathbf{d}\|}{v_{\text{path}}}$ is the segment duration based on the prescribed path speed $v_{\text{path}}$. The rotational motion is parameterized using axis-angle representation, where the rotation axis is 
\begin{equation}
\mathbf{k} = \frac{(-\mathbf{n}_i) \times (-\mathbf{n}_{i+1})}{\|(-\mathbf{n}_i) \times (-\mathbf{n}_{i+1})\|}
\end{equation}
and the rotation angle is 
\begin{equation}
\theta = \arctan2(\|(-\mathbf{n}_i) \times (-\mathbf{n}_{i+1})\|, (-\mathbf{n}_i) \cdot (-\mathbf{n}_{i+1}))
\end{equation}
yielding an angular velocity $\boldsymbol{\omega} = \frac{\mathbf{k}\theta}{t_{\text{segment}}}$. 

During baseline trials without the end-effector, the algorithm superimposes the admittance controller in Fig. \ref{fig:cont}C as the normal velocity with the translational velocity 
\begin{align}
\textbf{v}_\text{total} &= \textbf{v}_\text{trans} + \textbf{v}_\text{norm}\nonumber\\
\textbf{v}_\text{norm} &= v_\text{norm} \mathbf{n}_{\text{current}}\nonumber\\
v_\text{norm} &= c_1(F_m - F_{\text{target}}) + c_2 v_{\text{prev}}
\end{align}
to maintain constant contact force, where $c_1$ and $c_2$ are controller gains, $F_m$ is the measured normal force, $F_{\text{target}}$ is the desired contact force magnitude, and $v_\text{norm}$ clamped to $\pm v_{\text{max}}$. During trials with the end-effector, the admittance controller is turned off. Tracking trajectories of the orientations during the trials can be found in the supplementary materials Fig. S-1.

\subsection{\textit{Ex Vivo} Dynamic Motion Simulator Design}

This section describes the development of an actuated \textit{ex vivo} tissue platform that addresses a key limitation in robotic ultrasound testing: static phantoms cannot replicate the tissue motion from patient movements. We designed the platform to mimic the dynamic movements of a human abdomen (while supine) or retroperitoneal region (while prone) during respiration to achieve realistic validation.

The simulator has two broad components: a tissue staging platform, and an actuation system to simulate the body's movement during respiration. The tissue platform is designed to hold both \textit{ex vivo} tissue and tissue phantoms. An AK80-9 actuator drives a lead screw, actuating the tissue platform to move in an axial direction, following a similar lead screw mechanism as the system in \cite{chen_design_2024}. The simulator is shown in Fig. \ref{fig:diag}C. During testing, the platform oscillates in a sinusoidal pattern with an amplitude of 9.9 mm and a frequency of 14.6 cycles per minute, simulating human abdominal movements characteristic of quiet breathing in a supine position\cite{kaneko2012breathing}. 

To perform realistic experiments on the proposed end-effector, two types of materials were tested:  an \textit{ex vivo} porcine abdominal wall tissue purchased from a local supermarket and a tissue phantom. The porcine tissue comprised skin, subcutaneous fat, and underlying muscle layers.  For tissue phantom creation, $2.5\%$ high-acyl gellan gum (\%  mass) (VWR, PA, United States) was used following the procedure previously proposed \cite{prakash2023brain}. The developed phantom has Young's modulus greater than $25$ kPa, extrapolating values presented in \cite{prakash2023brain}.

\begin{figure*}[ht!]
    \centering
    \includegraphics[width=\textwidth]{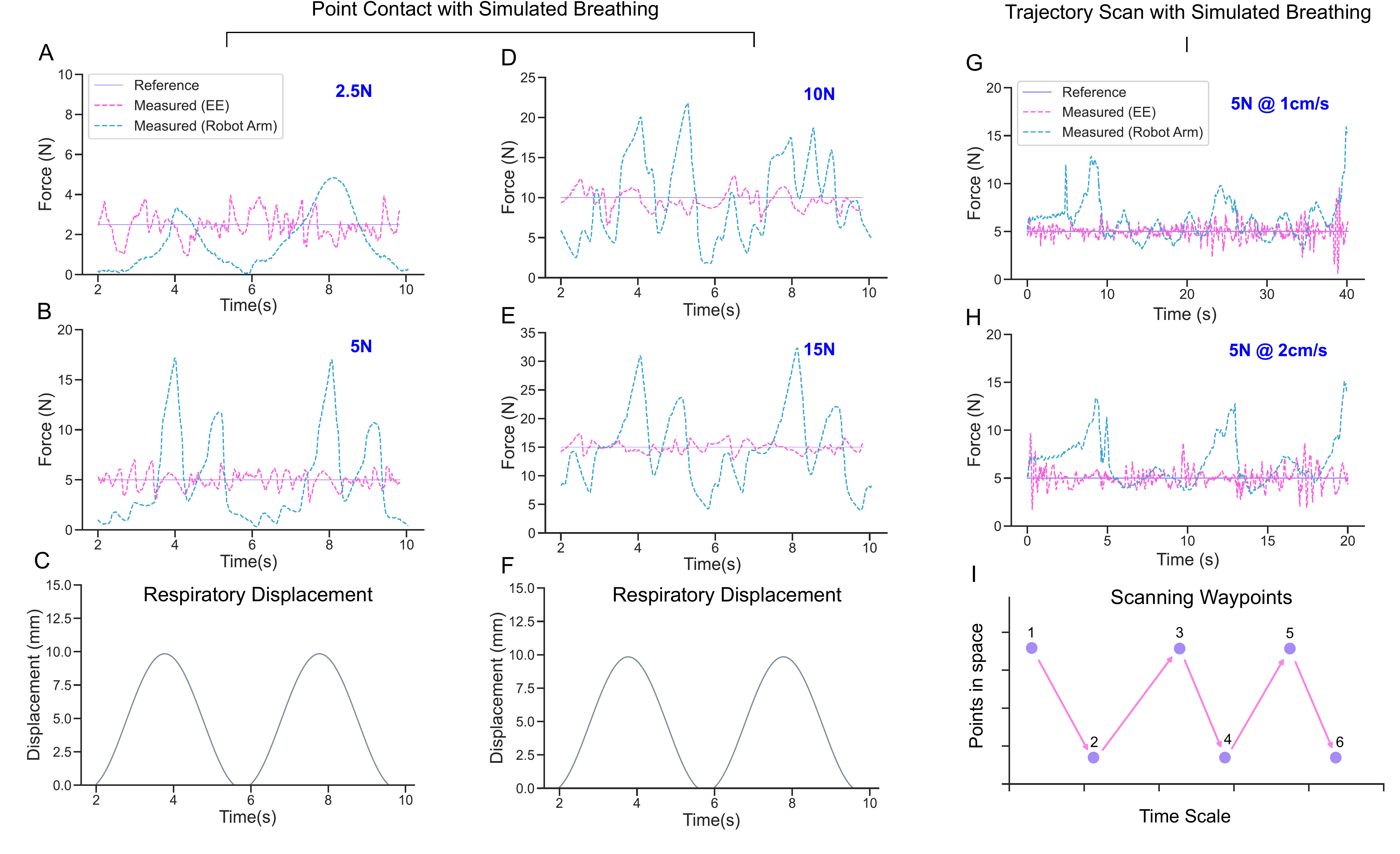} 
    \captionsetup{justification=justified}
    \caption{\small Experiment setup and force tracking results on porcine tissue with dynamic movement. A, B, D, E: force tracking plots of the end-effector (EE) and robot arm only conditions targeting 2.5 N, 5 N, 10 N, and 15 N during simulated breathing motion on point contact; C, F: the displacement caused by the dynamic motion simulator during the experiment; G, H: force tracking plots of the end-effector and robot arm only conditions targeting 5 N during simulated breathing motion on trajectory scan at 1 cm/s and 2 cm/s; I: indicates where waypoints occur along the scanning trajectories. The end-effector consistently shows superior force tracking performance with reduced overshoot and better target maintenance compared to the robot arm only configuration across all tested conditions.}
    \label{fig:comp}
    
\end{figure*}

\begin{figure}[h!]
    \centering
    \includegraphics[width=0.49\textwidth]{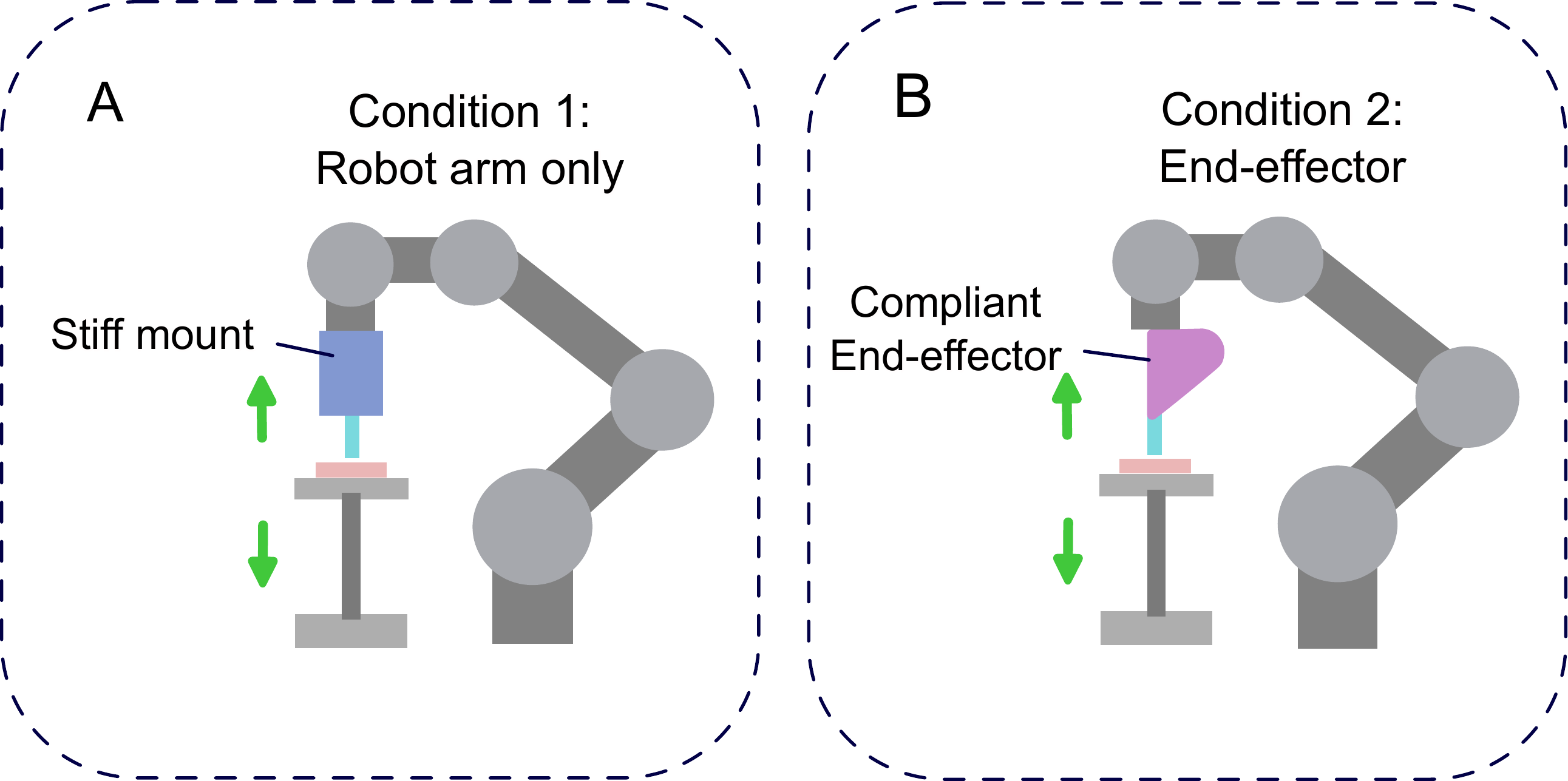} 
    \captionsetup{justification=justified}
    \caption{\small Experimental setup. A, Robot arm only baseline condition with rigid end-effector; B,  compliant end-effector condition with the end-effector attached to the robot arm. Both setups replicate realistic ultrasound scanning conditions for comprehensive system evaluation.}
    \label{fig:exp_conds}
\end{figure}

\begin{figure*}[h]
    \centering
    \includegraphics[width=\textwidth]{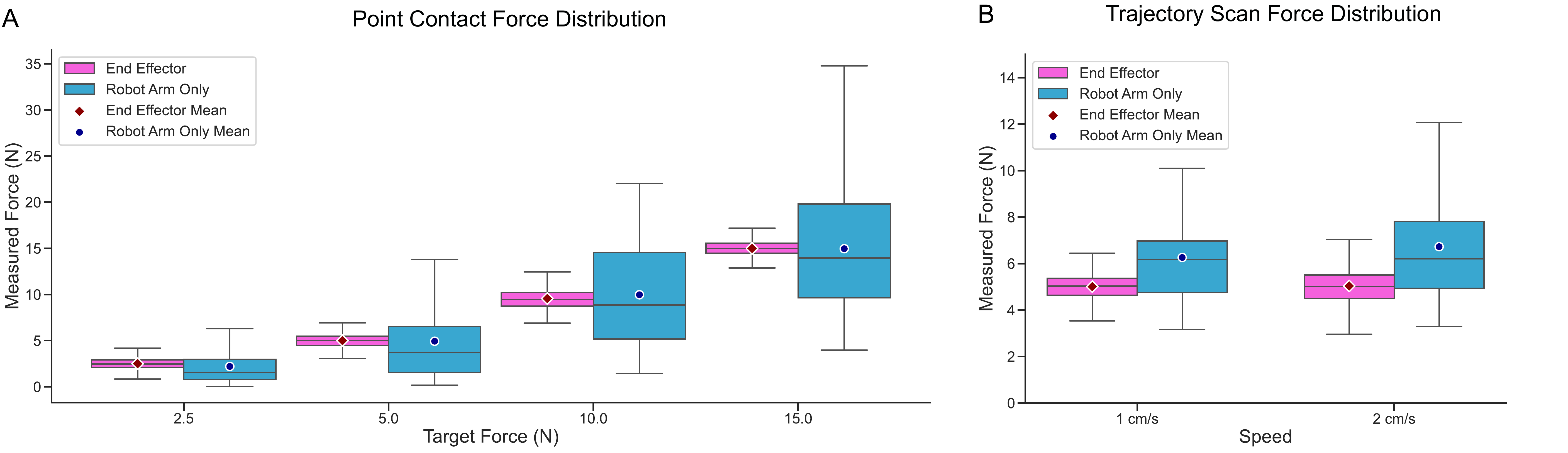} 
    \captionsetup{justification=justified}
    \caption{\small Force distribution box plots showing concatenated data from three measurements at each condition; A: point contact experiments at 2.5 N, 5 N, 10 N, and 15 N, with compliant end-effector and robot arm only conditions compared side by side; B: trajectory scan experiments at 1 cm/s and 2 cm/s, with compliant end-effector and robot arm only conditions compared side by side. The compliant end-effector consistently exhibits tighter force distributions with reduced variability and better centering around target values compared to the robot arm only configuration.}
    \label{fig:box}
    \vspace{-0.2cm}
\end{figure*}

\begin{figure*}[h]
    \centering
    \includegraphics[width=\textwidth]{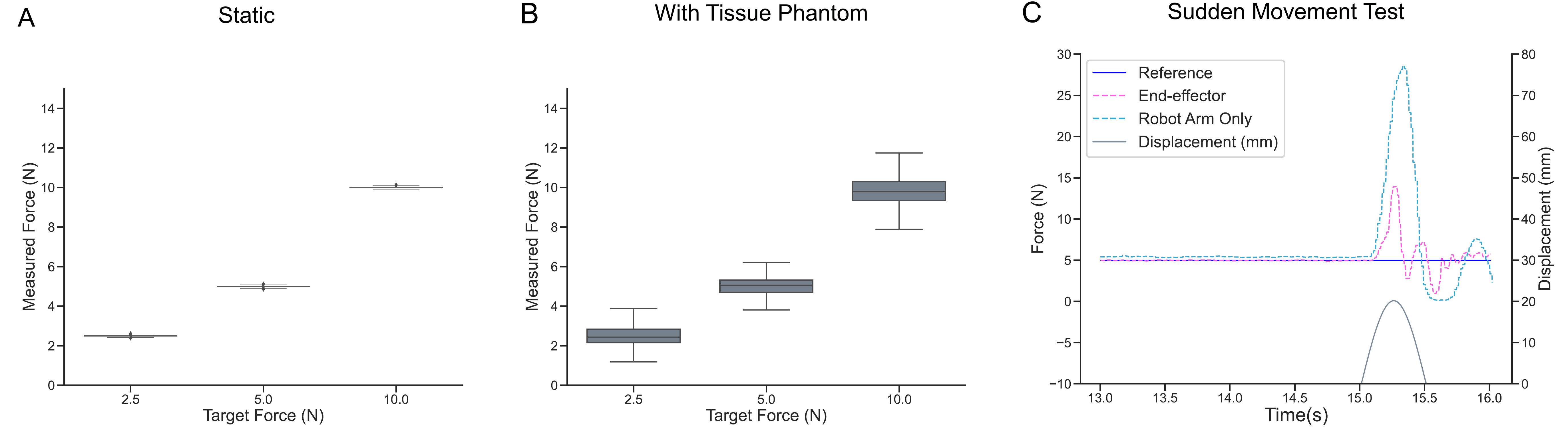} 
    \captionsetup{justification=justified}
    \caption{\small Force distribution boxplots and force tracking plot of supplementary experiments; A: force distribution of static point contact at 2.5 N, 5 N, and 10 N using the compliant end-effector; B: Force distribution of dynamic point contact at 2.5 N, 5 N, and 10 N on a tissue phantom using the compliant end-effector; C: force tracking comparison plots between compliant end-effector and robot arm only conditions under sudden movement while targeting 5 N. These supplementary experiments showed the end-effector's consistent performance across exploratory conditions, with clear superiority in handling sudden disturbances compared to the robot arm only configuration.}
    \label{fig:misc}
\end{figure*}

The point-contact experiments, which simulated abdominal ultrasound exams, used a flat surface to mount the tissue. In contrast, the trajectory-scan experiments were designed to emulate renal examinations, which not only involve scanning over a curved, uneven anatomical surface but also must contend with the rise and fall of the skin due to respiratory motion. We selected both scenarios because of their prevalence in clinical practice. A three-dimensional human male 3D model, reconstructed from medical imaging data\cite{Browne2023-no}, provided the basis for creating an anatomically accurate testing platform. A square section positioned directly above the right kidney on the posterior surface was extracted from this model and 3D printed using PETG plastic to replicate the natural topographical features of the human back. The printed platform incorporated a cubic cavity at its center, which was filled with Dragonskin medium silicone (Smooth-On Inc, PA, United States) to serve as both a reference object and target of interest during US imaging. Porcine abdominal wall tissue, cut to match the platform dimensions, was mounted over the surface with US coupling gel applied at the tissue-platform interface to ensure acoustic continuity. The tissue was then covered with fabric and scanned using an Intel RealSense SR305 depth camera (Intel, Santa Clara, CA, United States) to generate a three-dimensional point cloud representation of the surface topography. Finally, we selected six points in a zigzag pattern to cover an area roughly the size of an adult human kidney.

\section{Experiments and Results}

A series of force tracking experiments were done to evaluate the performance of our compliant end-effector while mounted on a UR3e robot arm (Universal Robots, Odense, Denmark) compared to the UR3e robot arm alone without the end-effector. The experimental setup is shown in Fig. \ref{fig:diag}C.

Two sets of experiments were conducted. The first set consisted of point contact experiments that simulate abdominal exams, where the contact point remains fixed and the goal is to maintain steady force while subjected to simulated respiratory motion (9.9 mm displacement, 14.6 breaths per minute \cite{kaneko2012breathing}) on the dynamic motion simulator. We used the robot arm alone as the baseline condition, seen in Fig. \ref{fig:exp_conds}. During the robot arm only conditions, the compliant end-effector was replaced with a rigid one, and the robot arm used the position-based PID controller from Fig. \ref{fig:cont}B. During the end-effector conditions, the robot arm was not powered, and the end-effector used the current-based PID controller from Fig. \ref{fig:cont}A. The tuned PID values used for each experiment can be found in Table~\ref{tab:pid}. Force measurements were obtained using a six-axis F/T 300 force sensor. Data collection involved recording force measurements at target forces of 2.5 N, 5 N, 10 N, and 15 N over 8-second intervals (two complete respiratory cycles), with n=3 measurements obtained for each condition. Mean values and root mean square error (RMSE) were calculated to assess force tracking accuracy. We defined forces above 150\% of the target force as ``excessive force", measuring this excessive force duration as a percentage of total operation time.

The second set of experiments added trajectory scanning in addition to the simulated respiratory motion on the dynamic motion simulator. Instead of maintaining a single contact point, the anatomically informed testing platform from the previous section was used as the surface to perform a trajectory scan. The robot arm followed a zigzag path covering a square area approximately the size of a male kidney. During the robot arm only conditions, the arm is fitted with a rigid end-effector, and used the admittance controller from Fig. \ref{fig:cont}C. During the end-effector conditions, the robot arm followed the scanning path without feedback-related adjustments, and the end-effector independently adjusted the contact force using the controller from Fig. \ref{fig:cont}A. All experiments were performed at 5 N, with path speeds of 1 cm/s and 2 cm/s compared to assess their effect on force tracking accuracy. Each condition was repeated n=3 times, with mean values and RMSE calculated to assess force tracking accuracy.

Three supplementary experiments were conducted as exploratory scenarios, distinct from the primary experimental protocol: (1) assessment of the compliant end-effector's positional accuracy on porcine tissue under static conditions (i.e., with dynamic motion disabled), (2) evaluation of the compliant end-effector's performance during dynamic motion using a tissue phantom as a substitute for biological tissue, and (3) comparative analysis of the compliant end-effector and the robotic arm alone in response to a sudden jerking movement induced by the tissue.

\subsection{Point Contact Experiments}
\subsubsection{Force Tracking on UR3e Robot Arm with PID controller}
The UR3e robot arm without the compliant end-effector served as the baseline condition, tested at force levels of 2.5 N, 5 N, 10 N, and 15 N while subjected to simulated respiratory motion. The measured mean forces, RMSE, and percentage of time above 150\% of the target force for each target force can be found in Table \ref{tab:force_tracking}. The large RMSE and high forces reflect the challenges of maintaining precise force control using conventional force control on the robot arm alone, especially in dynamic scenarios. See Fig.~\ref{fig:comp}(A, B, D, E) for the force tracking plots in a singular measurement, and Fig.~\ref{fig:box}A for the distribution of all measurements.
\begin{table}[h!]
\centering
\renewcommand{\arraystretch}{1.3}
\caption{Force tracking performance comparison under simulated respiratory motion}
\label{tab:force_tracking}
\begin{tabularx}{0.485\textwidth}{|X|c|c|c|c|}
\hline
\textbf{Force} & \textbf{2.5 N} & \textbf{5.0 N} & \textbf{10.0 N} & \textbf{15.0 N} \\
\hline
\hline
\multicolumn{5}{|c|}{\textbf{Robot Arm Only}} \\
\hline
Mean (N) & 2.20 & 4.94 & 9.97 & 14.96 \\
\hline
RMSE (N) & 2.09 & 4.30 & 5.43 & 6.99 \\
\hline
\% time above 150\% of target & 15.29 & 21.33 & 23.21 & 16.95 \\
\hline
\hline
\multicolumn{5}{|c|}{\textbf{w/ Compliant End-effector}} \\
\hline
Mean (N) & 2.50 & 5.01 & 9.56 & 14.99 \\
\hline
RMSE (N) & 0.61 & 0.77 & 1.09 & 0.86 \\
\hline
\% time above 150\% of target & 2.65 & 0.00 & 0.00 & 0.00 \\
\hline
\hline
\textbf{RMSE reduction (\%)} & 70.8 & 82.1 & 79.9 & 87.7 \\
\hline
\end{tabularx}
\end{table}

\begin{figure*}[ht!]
    \centering
    \includegraphics[width=\textwidth]{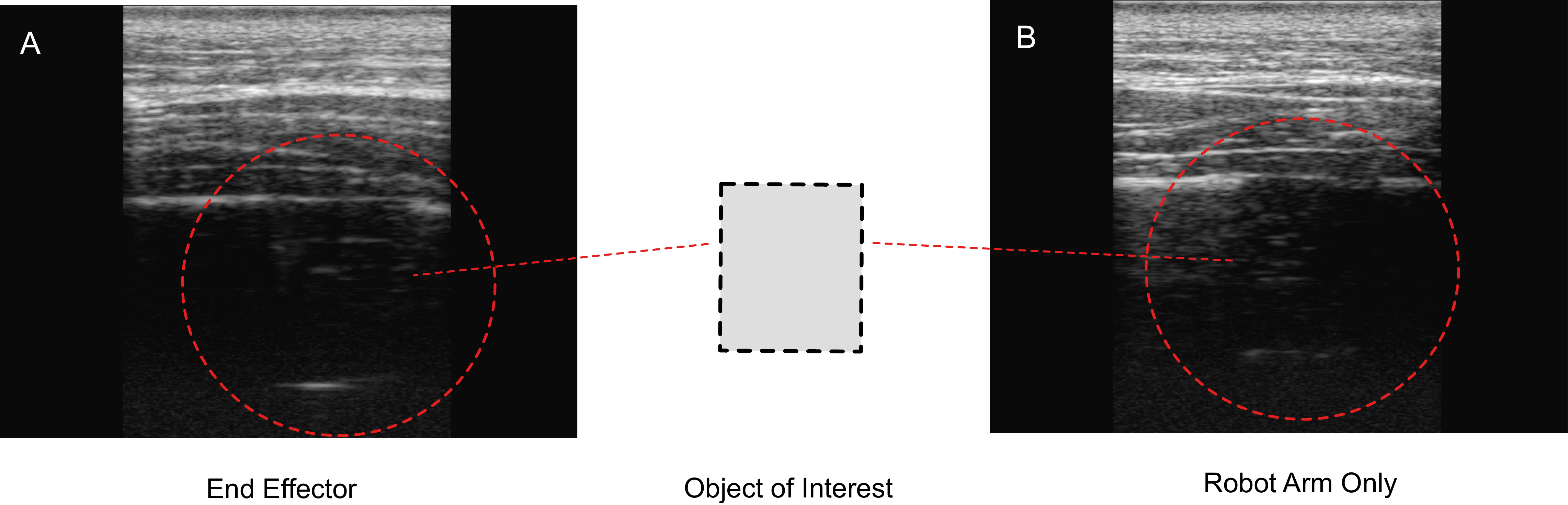} 
    \captionsetup{justification=justified}
    \caption{\small Representative US frames captured during trajectory scanning experiments showing the reference object (an embedded silicone cube beneath porcine tissue) for A: the compliant end-effector condition and B: the robot arm only condition. Both approaches successfully visualized the target silicone cube.}
    \label{fig:scan}
\end{figure*}

\subsubsection{Force Tracking on Compliant End-effector}
The end-effector was tested under the same simulated breathing conditions and target forces. The measured mean forces, RMSE, and percentage of time above 150\% of the target force for each target force can be found in Table \ref{tab:force_tracking}. This showed an average RMSE reduction of 80.1\% compared to the robot arm only condition across all four force levels. These results showed the system’s robustness in adjusting to dynamic axial motion throughout a wide range of forces.  See Fig.~\ref{fig:comp}(A, B, D, E) for the force tracking plots in a singular measurement and Fig.~\ref{fig:box}A for the distribution.

\subsection{Trajectory Scanning Experiments}

\subsubsection{Force Tracking on UR3e Robot Arm with Admittance Controller}
The UR3e robot arm without the compliant end-effector served as the baseline condition, tested at 5 N with path speeds of 1 cm/s and 2 cm/s. The robot arm followed a zigzag path and superimposed the admittance controller in Fig. \ref{fig:cont}C as the normal velocity perpendicular to the tissue surface to maintain constant contact force. The measured mean forces, RMSE, and percentage of time above 150\% of the target force for the respective path speeds can be found in Table \ref{tab:trajectory_force_tracking}. These results showed that under admittance control, the robot arm struggled to maintain the target mean force but achieved lower RMSE compared to the point contact experiment. Force tracking error increased by 23.2\% when path speed doubled. See Fig.~\ref{fig:comp}(G, H) for the force tracking plots in a singular trial and Fig.~\ref{fig:box}B for the distribution.

\subsubsection{Force Tracking on Compliant End-effector}
The compliant end-effector was tested during trajectory scanning, tested at 5 N with path speeds of 1 cm/s and 2 cm/s. The robot arm followed the path without feedback-based motion adjustments. The measured mean forces, RMSE, and percentage of time above 150\% of the target force for the respective path speeds can be found in Table \ref{tab:trajectory_force_tracking}. This showed an average RMSE reduction of 68.0\% compared to the robot arm only condition across both path speeds. See Fig.~\ref{fig:comp}(G, H) for the force tracking plots in a singular trial and Fig.~\ref{fig:box}B for the distribution. These results showed that the end-effector maintained a good performance on trajectory scanning. The RMSE at 1 cm/s is comparable to the point contact experiment, with force tracking error increasing by 21.1\% when path speed doubled.

\begin{table}[h!]
\centering
\renewcommand{\arraystretch}{1.3}
\caption{Force tracking performance comparison during trajectory scanning (5 N target)}
\label{tab:trajectory_force_tracking}
\begin{tabularx}{0.485\textwidth}{|X|c|c|}
\hline
\textbf{Path Speed} & \textbf{1 cm/s} & \textbf{2 cm/s} \\
\hline
\hline
\multicolumn{3}{|c|}{\textbf{Robot Arm Only}} \\
\hline
Mean (N) & 6.26 & 6.73 \\
\hline
RMSE (N) & 2.37 & 2.92 \\
\hline
\% time above 150\% of target & 5.13 & 10.03 \\
\hline
\hline
\multicolumn{3}{|c|}{\textbf{w/ Compliant End-effector}} \\
\hline
Mean (N) & 5.01 & 5.03 \\
\hline
RMSE (N) & 0.76 & 0.93 \\
\hline
\% time above 150\% of target & 0.00 & 0.00 \\
\hline
\hline
\textbf{RMSE reduction (\%)} & 67.9 & 68.2 \\
\hline
\end{tabularx}
\end{table}

\subsection{Supplementary Experiments}
\subsubsection{Force Tracking on End-effector with Porcine Tissue (Static)}
Force tracking on the end-effector was done on porcine tissue at target forces of 2.5 N, 5 N, and 10 N in a stable configuration, without axial motion. The system exhibited high precision in maintaining the desired force levels, with mean values of 2.50 N (RMSE: 0.03 N), 4.99 N (RMSE: 0.04 N), and 10.00 N (RMSE: 0.04 N), closely matching the target forces. These results confirm the system's ability to achieve excellent force tracking with effectively no steady state error under stable conditions. Refer to Fig.~\ref{fig:misc}A for the distribution.

\subsubsection{Force Tracking on End-effector with Tissue Phantom (Simulated Breathing Motion)}
The simulated breathing conditions were tested with a tissue phantom and the same target forces. The results showed measured mean forces of 2.50 N (RMSE: 0.59 N), 5.02 N (RMSE: 0.46 N), and 10.00 N (RMSE: 1.12 N). These results showed that the system achieved comparable performance on a homogeneous tissue phantom, suggesting that such phantoms may serve as viable alternatives to \textit{ex vivo} tissue for preliminary testing, thereby reducing the logistical challenges of tissue procurement. Refer to Fig.~\ref{fig:misc}B for the distribution.

\subsubsection{Force Tracking Comparison: UR3e Robot Arm and End-effector with Porcine Tissue (Sudden Movement)}
Finally, force tracking was assessed in the event of a sudden 20 mm sinusoidal displacement over 0.25 s rise time and 0.25 s fall time, both the robot arm and the compliant end-effector were tasked with maintaining a 5 N contact force on porcine tissue. The end-effector showed superior recovery and less peak force, with contact forces ranging from 0.96 N to 13.94 N, while the UR3e arm exhibited more pronounced fluctuations, with contact forces ranging from 0.12 N to 28.57 N. These results showed the end-effector's superior ability to maintain stable contact while not applying excessive force under abrupt motion. See Fig.~\ref{fig:misc}C for the comparative tracking plot.

\subsection{Imaging Analysis}
Although this study focused on safe force application and compliance without incorporating US imaging as control feedback, US data was collected for observational purposes to provide a more comprehensive evaluation. Despite slight differences in probe alignment between the rigid mount and compliant end-effector due to physical configuration differences, both systems successfully captured the silicone cube reference object embedded in the platform during scanning, as shown in Fig. \ref{fig:scan}. 

\begin{table}[h!]
\centering
\caption{Statistical comparison of image quality metrics across methods and experimental conditions}

\begin{tabularx}{0.5\textwidth}{l c c c c c c}

\toprule
& \multicolumn{3}{c}{\textbf{Method}} & \multicolumn{3}{c}{\textbf{Speed}} \\
\cline{2-4} \cline{5-7}
\textbf{Metric} & \textbf{Best} & \textbf{p-value} & \textbf{$\eta^2$} & \textbf{Best} & \textbf{p-value} & \textbf{$\eta^2$} \\
\hline
gCNR Mean$^{\uparrow}$ & EE & $<0.001$ & 0.088 & 1 cm/s & $<0.001$ & 0.018 \\
gCNR Std$^{\downarrow}$ & EE & $<0.001$ & 0.020 & 2 cm/s & $<0.001$ & 0.004 \\
SNR$^{\uparrow}$ & EE & $<0.001$ & 0.035 & --- & 0.336 & 0.000 \\
Entropy$^{\uparrow}$ & EE & $<0.001$ & 0.081 & 1 cm/s & $<0.001$ & 0.084 \\
SSIM$^{\uparrow}$ & RO & $<0.001$ & 0.016 & 1 cm/s & $<0.001$ & 0.045 \\
Frame Diff$^{\downarrow}$ & RO & $<0.001$ & 0.007 & 1 cm/s & $<0.001$ & 0.015 \\
\bottomrule
\end{tabularx}
\label{tab:results}
\footnotesize

\vspace{0.1cm}

EE: end-effector; RO: robot arm\\
$^{\uparrow}$ Higher is better, $^{\downarrow}$ Lower is better\\
--- = Not significant (p $>0.05$)

\end{table}

\begin{figure}[h!]
    \centering
    \includegraphics[width=0.5\textwidth]{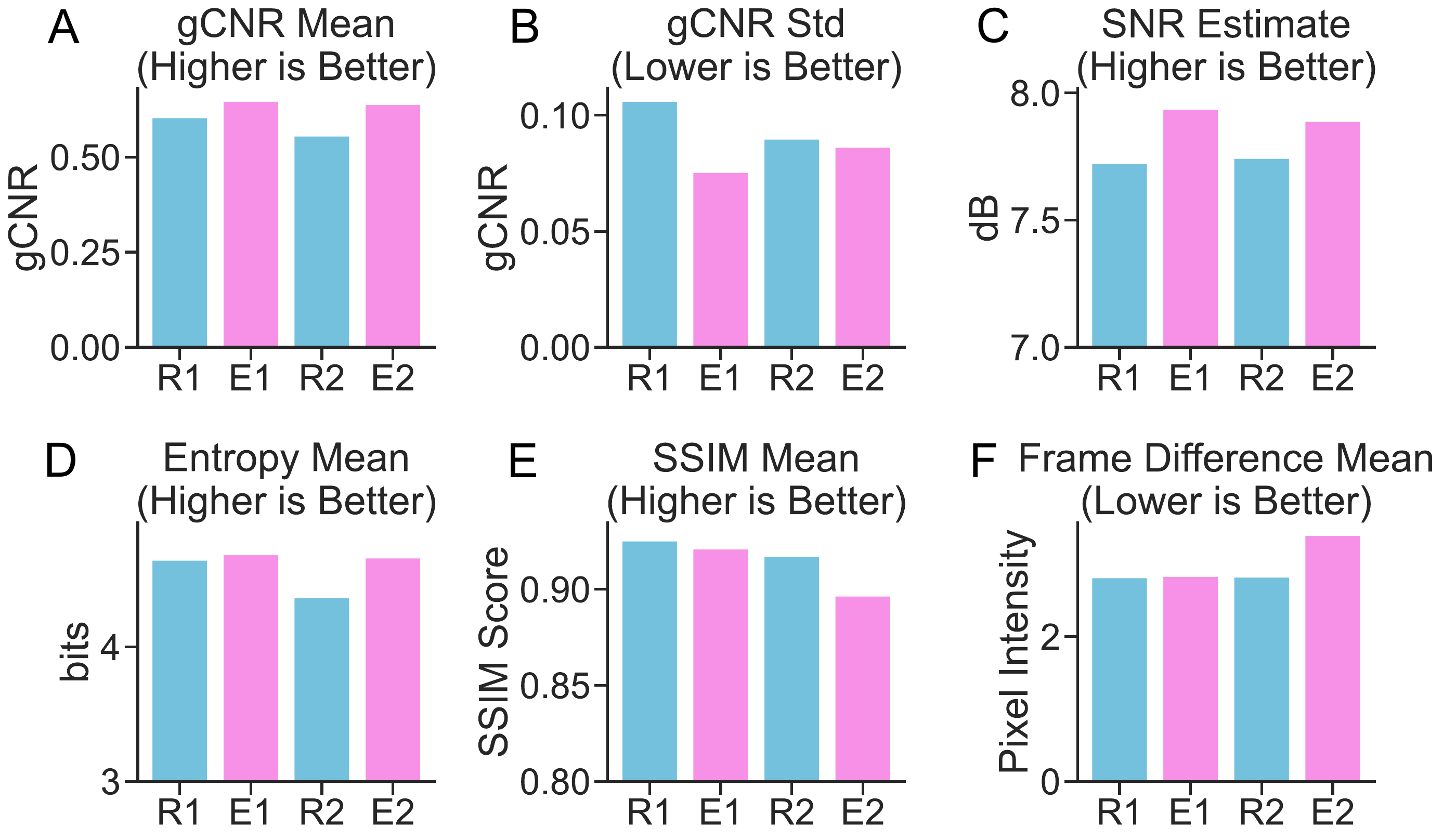} 
    \captionsetup{justification=justified}
    \caption{\small Image quality and stability metrics including generalized contrast-to-noise ratio (gCNR) A: mean and B: standard deviation; C: signal-to-noise ratio (SNR); D: entropy; E: structural similarity index measure (SSIM); and F: frame difference. R1: robot arm only condition at 1 cm/s; E1: compliant end-effector at 1 cm/s; R2: robot arm only condition at 2 cm/s; E2: compliant end-effector at 2 cm/s. The compliant end-effector shows statistically superior performance in four out of six image quality metrics, indicating improved imaging quality and consistency during trajectory scanning.}
    \label{fig:im}
\end{figure}

As visualized in Fig. \ref{fig:im}, quantitative analysis was performed using six metrics: generalized contrast-to-noise ratio (gCNR) mean and standard deviation, signal-to-noise ratio (SNR), entropy, structural similarity index measure (SSIM), and frame difference. Three metrics (gCNR, SNR, and entropy) assessed image quality using mean values. SSIM and frame difference quantified stability by measuring inter-frame visual content stability, while gCNR standard deviation evaluated quality consistency. Statistical comparisons (Table~\ref{tab:results}) were conducted using factorial ANOVA with Type II sum of squares to assess main effects of method (end-effector vs robot arm only) and speed (1 cm/s vs.\ 2 cm/s) and their interaction, with effect sizes reported as $\eta^2$, variance equality on gCNR standard deviation evaluated via Levene’s test, and pairwise differences examined using independent $t$-tests and Mann-Whitney U tests. Overall, the results showed that the end-effector outperformed the robot arm across four of the six metrics, and slower scan speeds (1 cm/s) yielded superior quality and stability compared to faster scans (2 cm/s). 
To qualitatively assess image quality, a clinician investigator performed a blinded evaluation of five paired image sets, each containing one image from each acquisition method randomly labeled A or B. The clinician rated three of the five pairs as having equivalent visualization quality, while the remaining two pairs each favored a different condition. These results suggest comparable image quality between the two conditions.

\section{Discussion}

The experimental results demonstrated superior force tracking accuracy of the end-effector during dynamic motion compared to conventional active compliance systems,  with a 0.83 N RMSE across tested forces compared to 4.70 N on the conventional setup or the 2.47 N variability of manual scans \cite{tsumura_robotic_2020}. Additionally, the dynamic tissue platform provided realistic surfaces and motion simulation for comprehensive evaluation. Force tracking comparisons utilized controlled baselines over similar published systems, as varying experimental conditions across studies make direct comparisons challenging. Nevertheless, the consistent improvements suggest significant practical benefits for robotic ultrasound scanning.

In the image quality metrics, the compliant end-effector showed superior performance in most quality metrics compared to the robot arm. While some variability in stability metrics was observed due to the compliant mechanism's quicker adaptive behavior, this represents a design trade-off prioritizing safety over rigid contact. Importantly, in a blinded qualitative assessment, a qualified clinician investigator found no significant differences in the images. This finding shows that our safety-focused approach achieves quantitatively superior performance while maintaining comparable image quality. 

A limitation of the current system is the use of PLA plastic in constructing the end-effector, which restricted the maximum force to 15 N, which is below the maximum clinical requirement of 17.3 N. The PLA structure's deformation and resulting belt slippage above 15 N prevent operation in the 15 - 20 N range. While we intended the system to provide controlled slippage at high forces as a safety feature (since extrapolated clinical data indicates forces exceeding 20 N could cause patient discomfort or pain\cite{waller2025pressure}), the 15 N mark falls short. Future iterations will address this limitation by using stiffer materials such as carbon-fiber-reinforced nylon or aluminum, allowing operation closer to the 17.3 N clinical maximum while maintaining protective slippage before reaching the 20 N discomfort threshold. Finally, while the results demonstrate consistent trends across repetitions (N=3), future work should include a larger number 
of trials to strengthen statistical robustness.

Future work should investigate different tissue types, movements, and scanning patterns through more comprehensive testing, ultimately progressing to in vivo patient trials. This study used simple model-free controllers to establish a baseline comparison between the end-effector and robot arm alone. Future work can experimentally validate and implement model-informed controllers using the dynamic model from Eq. 13-15, and explore integrating the robot arm's active compliance with the end-effector's control, rather than controlling them independently as done here to isolate effects. Another direction involves implementing variable force profiles rather than maintaining constant values. Finally, although this work focused exclusively on force feedback for safety and compliance, future studies could incorporate US image feedback into the control scheme. This integration would enable real-time adjustment of scanning parameters based on image quality metrics, automatic tracking of anatomical features during respiratory motion, and optimization of probe orientation to maintain optimal acoustic windows throughout the examination.

From a clinical perspective, the end-effector's ability to maintain accurate force tracking during dynamic movement is significantly better than using conventional active compliance control on only the robot arm. The precise force tracking could lead to more consistent US images, enhancing diagnostic accuracy and reducing the need for repeat scans. Additionally, the end-effector's rapid response to sudden movements (peak force of 13.94 N compared to 28.57 N for the robot arm) suggests a reduced risk of patient discomfort or injury during unexpected patient movements, which is important in pediatric or geriatric care where compliance may be important. 

\section{Conclusion}

The compliant QDD end-effector presented in this paper addresses key compliance challenges in robotic US imaging by ensuring patient safety and comfort through passive mechanical compliance while maintaining precise force control. The experiments validate its ability to achieve stable force application during simulated dynamic conditions that mimic respiratory motion, which is essential for abdominal and retroperitoneal imaging.
These strengths make the proposed system well-suited for potential clinical applications and improves safety and reliability in robotic ultrasound.

\section*{Acknowledgments}
The authors would like to thank George Delagrammatikas, Ethan LoCiero, Amy Strong, Kent Yamamoto, and Chad Miller for providing valuable feedback, as well as the Duke Innovation Co-lab for supporting this project.

\bibliographystyle{IEEEtran}
\bibliography{bib}

\end{document}